# Automated Blood Cell Detection and Counting via Deep Learning for Microfluidic Point-of-Care Medical Devices

**Tiancheng Xia[1], Richard Jiang[1], YongQing Fu[2] and Nanlin Jin[3]**

[1]School of Computing & Communication, Lancaster University, Lancaster, UK, LA1 4WY
[2]Faculty of Engineering & Enviornment, Northumbria University, Newcastle, UK, NE1 8ST
[3]Computer and Information Sciences, Northumbria University, Newcastle, UK, NE1 8ST

Correspondent e-mail: r.jiang2@lancaster.ac.uk

**Abstract.** Automated *in-vitro* cell detection and counting have been a key theme for artificial and intelligent biological analysis such as biopsy, drug analysis and decease diagnosis. Along with the rapid development of microfluidics and lab-on-chip technologies, *in-vitro* live cell analysis has been one of the critical tasks for both research and industry communities. However, it is a great challenge to obtain and then predict the precise information of live cells from numerous microscopic videos and images. In this paper, we investigated *in-vitro* detection of white blood cells using deep neural networks, and discussed how state-of-the-art machine learning techniques could fulfil the needs of medical diagnosis. The approach we used in this study was based on Faster Region-based Convolutional Neural Networks (Faster RCNNs), and a transfer learning process was applied to apply this technique to the microscopic detection of blood cells. Our experimental results demonstrated that fast and efficient analysis of blood cells via automated microscopic imaging can achieve much better accuracy and faster speed than the conventionally applied methods, implying a promising future of this technology to be applied to the microfluidic point-of-care medical devices.

## 1. Introduction

Microfluidic technologies [1, 2] have recently found wide-range applications in biological and medical applications, such as lab-on-chip and point-of-care (POC) diagnostic devices, which revolutionised the personalized medicine and rapid disease diagnosis. Point-of-care testing (or bedside testing) is generally defined as medical diagnostic testing at or near the point of caor in other words, at the time and place of patient care. This contrasts with the conventional treatment, in which testing was wholly or mostly confined to the medical laboratory. In this case, the specimens are often taken away from the point of care and then hours even days will be waited for the results, during which the point of care is asked to wait before the critical information is obtained.

Such the POC diagnosis has facilitated a paradigm shift from therapeutic treatments to predictive, personalized and preventive ones [1, 2], and from the conventional diagnostic tests performed inside the clinical labs settings to near-patient ones. This will enable doctors to have timely diagnostic information to make quick decisions regarding to further diagnosis or immediate treatments. At the same time, patients are hugely benefited from the ease of usages of the POC devices, which allow them to personally monitor their own health in reliable and quantified ways, simply being at home. There is no requirement for the tested samples to be delivered to a lab and no need for the results to be transmitted physically manually or over long distance and time. Doctors, nurses, or even patients could perform the tests and immediately receives the results on the spot, thereby, saving huge amount of time with the rapid diagnosis.

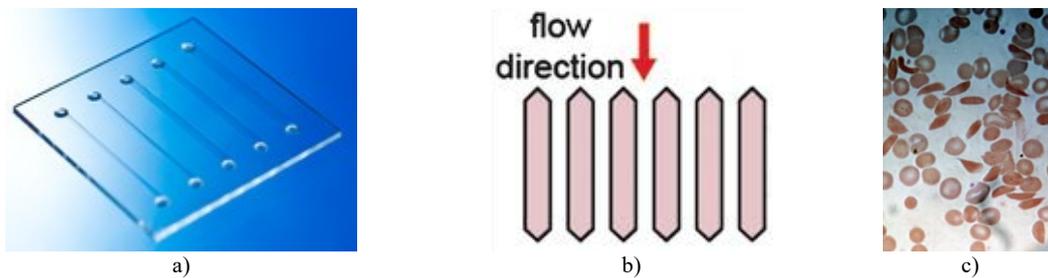

a) b) c)

Figure 1. Microfluidic point-of-care diagnostics of blood cells. a) Microfluidic device; b) Blood flow illustration; c) Microscopic inspection of blood cells.

Among various lab-on-chip based diagnosis tasks [1, 2], microfluidics based blood cell diagnostics as shown in Fig. 1 is one of critical ones. White blood cell (WBC) counting, a routine test to measure the number of white blood cells for a patient, is an important means for medical diagnosis. A low WBC count may be linked to a toxic reaction, a viral infection, or side-effect from chemotherapy, or a disease in the bone marrow which limits the body's ability to produce the normal WBCs, whereas a high WBC count might be linked to an imperative sign of an infection or leukaemia.

In hospitals, the conventional diagnostic procedures for blood cell analysis involve microscopic inspection of peripheral blood samples. For decades, this operation is performed by experienced operators with two main analysis procedures: classification and counting of the cells. However, lacking of automation and intelligent procedures has become a critical barrier for the integration of microscopic image analysis into microfluidic POC diagnostic system [3, 4]. So far, blood cell counting in the commonly used POC devices is mostly done using flow cytometry, which is based on monitoring changes of the fluorescence signals as a quantitative tool [4], however, this is far inferior and less accurate compared to cell-by-cell counting based on the microscopic images. As it is well-known, microscopic cell counting is non-invasive and free of fluorescence dyes, and it is based on the lab-on-chip devices and hence could drastically reduce the cost of the POC diagnostics.

In this work, aiming at the integration of automated microscopic image analysis into microfluidic POC device for blood cell counting, we investigated the methods of live cell detection techniques based on a recently developed artificial intelligence (AI) approach. Live-cell imaging experiments [3-6] provide the possibilities of identification, tracking, and analysis of cells from the microscopic images using a computer vision-based AI method. These experiments have been applied for wide-range applications such as biomedicine, material science and organic chemistry. Automated tracking of cell populations *in-vitro* using time-lapse microscopy images helps high-throughput spatiotemporal measurements of a wide variety of cell behaviours, including migration (translocation), mitosis (division), quiescence (inactivity) and apoptosis (death) of cells in addition to the reconstruction of cell lineages (mother–daughter relations) [7]. These capabilities are of enormous values in many areas of biomedical engineering, such as stem cell research, oncological studies, tissue engineering, drug discovery, genomics, and proteomics [7, 8].

Early live cell detection and tracking methods conventionally apply classic computer vision techniques such as segmentation algorithms, motion detection, level-set methods, image descriptors (such as SIFT/LBP/HOG). However, fully automatic cell tracking faces many challenges [6-10] including poor contrasts with high noise levels, irregular cell contours, difficulty for entry and exit of the cells, all of which make it difficult to deploy to commercial applications such as microfluidic POC devices. On the other hand, classic machine learning approaches such as subspace learning [11-12], ensemble approaches [13-14] and random forests [15] may work well on pattern analysis tasks.

However, in medical diagnosis, patients and users may demand an extremely high accuracy to avoid any medical misdiagnosis that could cause the fatal results. Hence, it is extremely challenging when AI is applied for medical care, which is a bit tricky in the sense that any misclassification will take legal or ethical consequences, and often result in a huge loss to compensate the victim of the misdiagnosis.

Recently, deep neural networks have been witnessed with a huge success in many challenging pattern analysis tasks [16-18]. Particularly for image analysis, convolutional neural networks (CNN) [19-20] has been used for the detection of in-vitro stem cells and achieved a considerably high accuracy (around 90%). In this paper, targeting at exploiting this technique for microfluidic blood cell analysis, we will examine the potentials of the recently-proposed object detection methods and explore its potential for fully-automated microfluidic POC technology. In this work, we select a method called Faster Region-based CNN (Faster-RCNN) for white blood cell examination, and investigate its robustness for potential commercial usages.

## 2. Faster RCNNs based Object Detection

In the past five years, deep learning based object detection has been evolving into a series of different methods, including different versions called R-CNN [21], Fast R-CNN [22] and Faster R-CNN [23], in an order with the evolution of objection detection. Region proposal networks (RPNs) are critical in all these types of approaches. R-CNN and Fast R-CNN firstly estimate the possible region proposals using a method called selective search (SS) [20]. A CNN-based network is then employed to classify the object and also detect its bounding box. The main dissimilarity between these two is that R-CNNs feed the region proposals at a pixel level into CNN for detection while the Fast R-CNN based networks take the region proposals at a feature map level. In both R-CNN and Fast R-CNN, the region proposal unit (i.e. SS) and the detection network (classifier) are decoupled. Obviously, such decoupling is not a best approach. For instance, the SS may have a false negative value, and this error will then pass over to the detection network successively. Apparently, it could be better to couple both stages with each other and associate them together. In the Faster R-CNN, RPNs using SS [20] is replaced by RPN using CNN, and this CNN is shared with detection network, and hence SS is removed from its cascaded process.

### 2.1. Region Proposal network

In object detection using Faster RCNN, RPN is the critical component and has been proven very efficient. Its purpose is to generate multiple potential region candidates identified within a given image. RPN produces the proposals for the objects, and owns a specialized architecture itself. Here, we can further

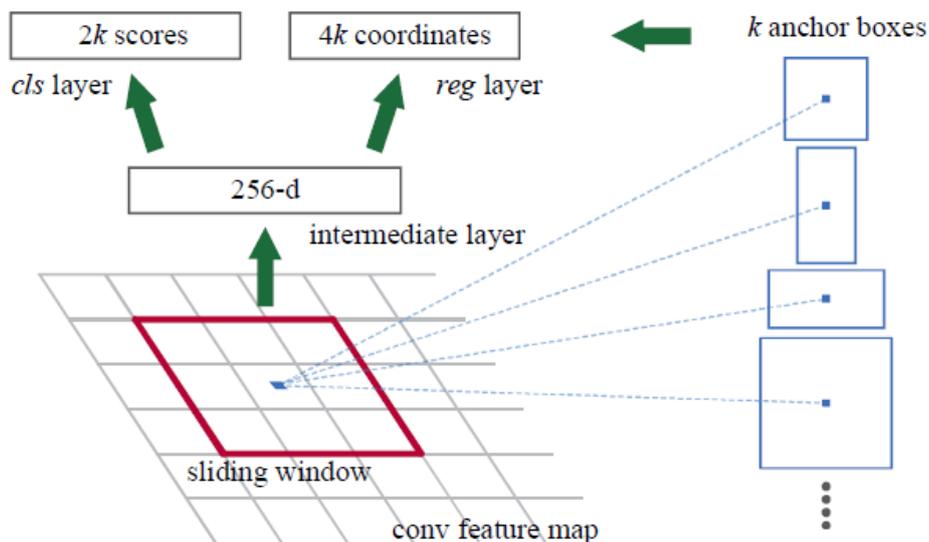

Figure 2. The typical RPN architecture [23].

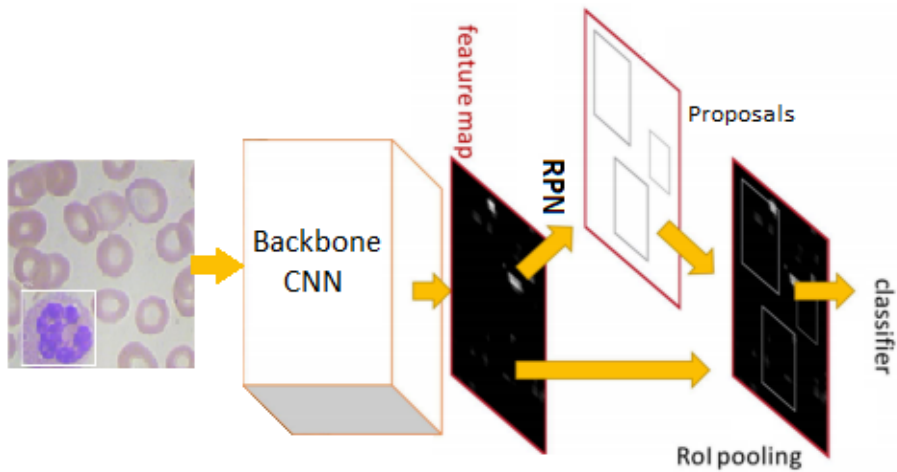

Figure 3. Overall architecture of Faster R-CNN detection network.

breakdown the RPN architecture into different components.

As shown in Fig. 2, RPN has a classifier as well as a regressor. Here, we bring in the concept of anchors, the central point of the sliding window. In the AlexNet, the dimension of the sliding window is 256-dim; in the VGG-16, it is 512-dim. The classifier calculates the probability of a proposed region having the expected object. The regressor regresses the coordinates of the region to make them fit better with the objects.

For any image, there are two important parameters: aspect-ratio and scale. Here, aspect ratio is the width of image divided by the height of image, and scale is the size of the image. If we choose 3 scales and 3 aspect-ratios, we will then have total of 9 proposals possible for each pixel. This is the way to estimate the value of $k$, and for this case, $k=9$. For the whole image, number of anchors is then counted as $W*H*k$.

This region proposal algorithm is robust to translations, with a merit of the translational invariance. It has multi-scale anchors. Instead of "*Pyramid of Filters*", this algorithm has "*Pyramid of Anchors*". As a result, this is more cost effective than previously proposed algorithms such as Fast RCNN. These anchors are assigned labels based on two factors: 1) The anchors with highest Intersection-over-Union (IoU) overlap with a ground truth box; 2) the anchors with IoU overlap higher than 0.7.

As it is expected, the RPN is an algorithm that needs to be trained, with regards to its Loss Function:

$$L(\{p_i\}, \{t_i\}) = \frac{1}{N_{cls}} \sum_i L_{cls}(p_i, p_i^*) + \lambda \frac{1}{N_{reg}} \sum_i p_i^* L_{reg}(t_i, t_i^*) \qquad (1)$$

The first term is the classification loss over two classes (object or non-object). The second term denotes how much the regression loss of bounding boxes is when there is an object (i.e. $p_i^* = 1$) in that region. Therefore, RPN networks will check over the locations that contain objects. The corresponding locations and bounding boxes will then be handed over to the detection network to identify which class the object is and what bounding box that object has. While the regions could be overlapped with each other, a strategy called non-maximum suppression (NMS) is then used to reduce the number of proposals, typically from more than 6000 to $N$ ($N=300$).

## 2.2. Detection Network

Next to the RPN, the following part in Faster R-CNN is illustrated in Fig. 3. First of all, ROI pooling is carried out. The pooled area is then passed through the CNN layers and two fully-connected branches as the class softmax and the bounding box regressor, respectively. Since the CNN backbone layers are shared in extracting the feature maps with different outputs at the end, training procedure is very different from the ones for RCNN and Fast RCNN. The training steps of Faster RCNN include,

1) Re-train (fine-tune) RPN with the pre-trained model (such as AlexNet);

Table 1. The test results on the WBC dataset

| No of Test Images | False Positive | False Negative | Overall Miss Rate | Overall Accuracy |
|---|---|---|---|---|
| 314 | 1 | 4 | 1.3% | 98.4% |

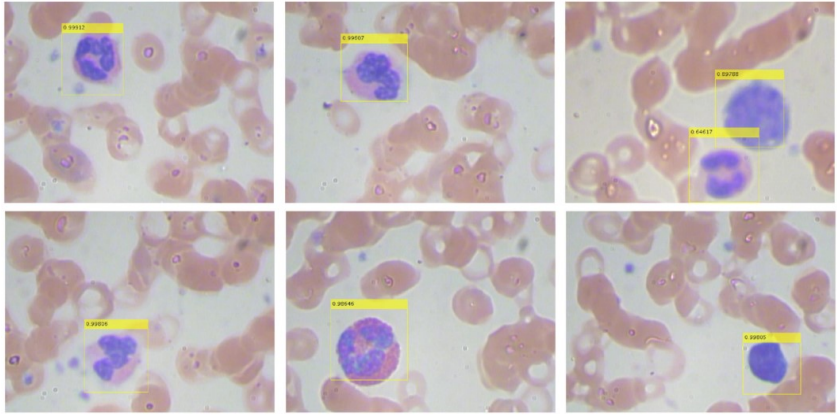

Figure 4. In-vitro white blood cell detection.

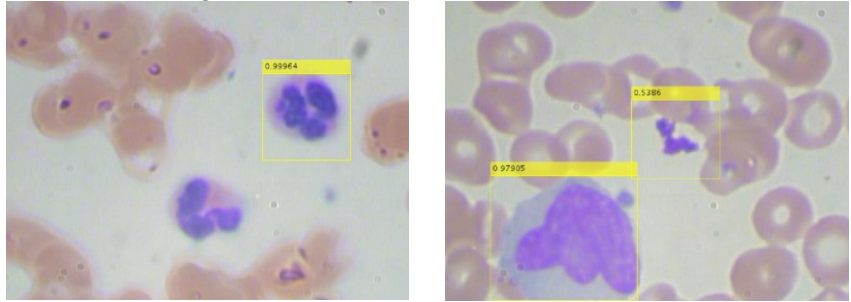

Figure 5. The false negative (left) and false posiive (right) test images

2) Re-train (fine-tune) a separate detection network with imagenet pre-trained model;
3) Use the detector network to initialize RPN training, while fixing the shared CNN layers and only tuning unique layers of RPN;
4) Keep the CNN layers fixed, fine-tune the unique layers of the detector network.

After the above steps, the object detector is ready to use on the trained classes of objects.

## 3. Using Faster RCNN for Blood Cell Detector

### 3.1. Experimental Setup

While mostly Faster RCNN detector has been trained with large datasets such as imagenet, it can be re-trained for specific tasks via a mechanism called transfer learning, as detailed in Section 2. In this work, we aim to examine the performance of Faster RCNN detector for our microfluidic blood cell analysis. Obviously, blood cells are not a class in the original imagenet classes, and thus we will need to re-train the detector so that it could be used to detect blood cells. To enable this, we used a small dataset of white blood cells for the transfer learning, freezed the backbone layers and re-trained the parameters of the FC layers for two classes, namely white blood cell and other objects. As shown in Fig. 3, an RPN is trained to quickly identify all WBC-like regions, and then a cascaded classifier is trained to determine if the detected region contains a WBC.

In our experiment, we analysed a small dataset consisting of 364 images [24], which were split into two subsets: 50 images for training and 314 images for testing. In the training setup, the learning rate was set as $1\times10^{-4}$ for the RPN and $1\times10^{-5}$ for the CNN. The maximum number of training epochs was set to be 15 for all the stages. The input layer of neurons was set to be $32\times32\times3$.

*3.2. Experimental Results*

After our WBC detector was trained on the small train dataset, we run the test on the rest of images, e.g., 314 test images. Fig. 4 shows the detection results from our re-trained detector on the tested images, while the detected WBCs are shown in the yellow rectangle with their probability specified in the yellow labels. Table 1 lists the statistic results of the experiment using the test dataset. Among all the 314 test images, we have only four WBCs missed in the detection (namely False Negative, FN), and one detection is wrongly selected as positive (False Positive, FP). In overall, the miss rate is only 1.3% and the detection accuracy is around 98.4%. Fig. 5 shows the FN and FP detection cases, where in the left image a white blood cell was missed in the detection and in the right image a wrong region was misclassified as a white blood cell.

**4. Further Discussion**

Such a high detection accuracy achieved by our CNN detector implies the aoolicability to use this technique for the real-life medical purpose. However, test results on the well-selected samples may still yet to be limited. More vigorous tests are needed to establish sufficient confidence for investors to support any real business-targeted development.

Mostly, to guarantee a reliable or acceptable accuracy for medical use, the accuracy under any conditions needs to be higher than 90% at a relatively fast speed. It may still have a long way to go before the real-life applications. Just like face recognition, test accuracy on datasets can easily be boosted to 99% but the real life test in the wild can have a low accuracy less than 20%, as reported recently by London police. However, our research may throw lights on its further exploitation for microfluidic POC device design.

**5. Conclusion**

In this work, we investigated the fully automatic WBC counting method based on microscopy image analysis using deep learning and validated the accuracy of the CNN based method in experiments. While it is a new technique to count blood cells in microfluidic devices for the POC service, our work shows a new pathway from image analysis that can be potentially applied for POC devices, though it may still be a long way to go to enable the pragmatic use in clinic practice.

**[References]**
[1] Chin, C. D., *et al*, 2012, Commercialization of microfluidic point-of-care diagnostic devices, *Lab Chip*, 2(12), p.2118-2134.
[2] Quesada-Gonzále D and Merkoçi A, 2018, Nanomaterial-based devices for point-of-care diagnostic applications. *Chemical Society Reviews*. 47 (13), p.4697–4709.
[3] Labati R. D., *et al*, 2011, ALL-IDB: the acute lymphoblastic leukemia image database for image processing, *Proc. of the 2011 IEEE Int. Conf. Image Processing*, Brussels, Belgium, p.2045-2048.
[4] Basiji D. A., *et al*, 2007, Cellular image analysis and imaging by flow cytometry, *Clin. Lab. Med.*, 27 (3), p.653–670.
[5] Mualla F., *et al*, 2013, Automatic cell detection in bright-field microscope images using SIFT, random forests, and hierarchical clustering, *IEEE Trans. Medical Imaging*, 32 (12). p.2274-2286.
[6] Jiang R., *et al*, 2010, Live-cell tracking using SIFT features in DIC microscopic videos, *IEEE Tran. Biomed Eng.*, 57, p.2219–2228.
[7] Harder, N., et al. 'Large-scale tracking and classification for automatic analysis of cell migration and proliferation, and experimental optimization of high-throughput screens of neuroblastoma cells', *Cytometry Part A*, (2015): 524-540.
[8] Armitage, J. P., 'Regulation of the expression and positioning of chemotaxis and motor proteins in Rhodobacter sphaeroides', *Diss. University of Oxford*, 2010.
[9] Tuysuzoglu, A., 'Robust inversion and detection techniques for improved imaging performance', *Diss. Boston University*, 2014.
[10] Fernyhough, E. N., 'Automated Segmentation of Structures Essential to Cell Movement', *Diss.*

*University of Leeds*, 2016.
[11] Jiang R, et al, 2010, Multimodal Biometric Human Recognition for Perceptual Human–Computer Interaction, *IEEE Trans. Systems, Man, & Cybernetics Part C*, 40(5), p.676.
[12] Jiang R, et al, 2010, Face recognition in global harmonic subspace, *IEEE Trans. Information Forensics and Security*, 5(3), p.416-424.
[13] Jiang R, et al, 2017, Emotion recognition from scrambled facial images via many graph embedding, *Pattern Recognition*, 67, p.245-251.
[14] Jiang R, et al, 2016, Face recognition in the scrambled domain via salience-aware ensembles of many kernels, *IEEE Trans. Information Forensics and Security*, 11(8), p.1807-1817.
[15] Jiang R, et al, 2016, Privacy-protected facial biometric verification via fuzzy forest learning, *IEEE Trans. Fuzzy Systems*, 24(4), p.779-790.
[16] Jiang Z, et al, 2019, Social Behavioral Phenotyping of Drosophila with a 2D-3D Hybrid CNN Framework, IEEE Access, p.100.
[17] Storey G, et al, 2018, Integrated Deep Model for Face Detection and Landmark Localisation from 'in the wild' Images, IEEE Access, p.200.
[18] Storey G, et al, 2017, Role for 2D image generated 3D face models in the rehabilitation of facial palsy, IET Healthcare Technology Letters.
[19] Wang, Y., *et al*, 'Stem cell motion-tracking by using deep neural networks with multi-output', *Neural Computing and Applications*, 2017, pp.1-13.
[20] Zitnick C L and Dollar P, 2014, Edge boxes: Locating object proposals from edges, European Conference on Computer Vision (ECCV), p.1.
[21] Girshick R, 2014, Rich feature hierarchies for accurate object detection and semantic segmentation, *IEEE Conference on Computer Vision and Pattern Recognition*, p.1.
[22] Girshick R, 2015, Fast R-CNN, *IEEE International Conference on Computer Vision*, p.1.
[23] Ren, S., He, K., Girshick, R., Sun, J.: Faster R-CNN: towards real-time object detection with region proposal networks. In: NIPS (2015)
[24] BCCD Dataset, http://github.com/akshaylamba/all_CELL_data, accessed by Feb 2018.